\documentclass[letterpaper, 10 pt, conference]{ieeeconf}
\usepackage{cite}
\usepackage{amsmath,amssymb,amsfonts}
\usepackage{algorithm, algorithmic}
\usepackage{graphicx}
\usepackage{textcomp}
\usepackage[c2 , nocomma]{optidef}
\usepackage{color}
\usepackage{mathtools}
\usepackage{lipsum}
\usepackage{eucal}
\usepackage{subcaption}
\usepackage{graphicx}
\def\BibTeX{{\rm B\kern-.05em{\sc i\kern-.025em b}\kern-.08em
    T\kern-.1667em\lower.7ex\hbox{E}\kern-.125emX}}
    
\IEEEoverridecommandlockouts 
\begin{document}

\title{\LARGE \bf An Active Learning Based Robot Kinematic Calibration Framework Using Gaussian Processes}

\author{Ersin Daş$^{1}$ and Joel W. Burdick$^{1}$
\thanks{*``This work was supported by NASA Grant 80NSSC21K1032.''}
\thanks{$^{1}$Ersin Daş and Joel Burdick are with the Department of Mechanical and Civil Engineering, California Institute of Technology, Pasadena, CA 91125, USA.
        {\tt\small $\{$ersindas,jburdick$\}$@caltech.edu}}
}

\maketitle
\pagestyle{plain}

\begin{abstract}
Future NASA lander missions to icy moons will require completely automated, accurate, and data efficient calibration methods for the robot manipulator arms that sample icy terrains in the lander's vicinity.
To support this need, this paper presents a Gaussian Process (GP) approach to the classical manipulator kinematic calibration process.  Instead of identifying a corrected set of Denavit-Hartenberg kinematic parameters, a set of GPs models the residual kinematic error of the arm over the workspace.  More importantly, this modeling framework allows a Gaussian Process Upper Confident Bound (GP-UCB) algorithm to efficiently and adaptively select the calibration's measurement points so as to minimize the number of experiments, and therefore minimize the time needed for recalibration.  The method is demonstrated in simulation on a simple 2-DOF arm, a 6 DOF arm whose geometry is a candidate for a future NASA mission, and a 7 DOF Barrett WAM arm. 
\end{abstract}


\section{Introduction}

{\em Kinematic Calibration (KC)} of robot manipulators is a classic problem \cite{hollerbach1989survey,khalil2002modeling,nubiola2013absolute}.  Calibration was originally motivated by the need for precise positioning of factory robot tools and end-effectors.  With limited real-time sensing, factory robots depend upon their open-loop positioning precision to reliably accomplish sophisticated tasks.  As further discussed below, there are several approaches to kinematic calibration, methods to deploy calibration in a variety of situations, and approaches to better automate the calibration process.

This paper revisits the calibration problem with new methods because of proposed NASA missions to the icy moons of Europa and Enceladus, which may host subsurface oceans.  Cryovolcanic activity might transport water/ice from a subsurface ocean through surface cracks.  As part of a Europa lander mission \cite{hand2022}, a robotic arm will gather surface material samples from the vicinity of these cracks, and analyze the local terrain using tools (scoops, drills, penetrometers, and bevometers).  The icy deposits may contain molecules that indicate the presence of life in the subsurface oceans.

These missions will likely be battery powered, which severely limits their duration. Round trip communication delays make remote manipulator operation impractical.  Hence, highly autonomous and efficient science operations are crucial.  While Europa is not a factory environment, kinematic accuracy of the sampling robot arm is still crucial because proper science interpretation of the samples requires accurate localization of the their origins.  Furthermore, since deep space missions must rigorously minimize mass, the sampling arm and its tools will likely be very lightweight, and thus more deformable than a factory robot.

Clearly, a data-efficient arm self-calibration process is indispensable for reliable icy moon sample collection and high mission efficiency.  At a minimum, upon landing and first deployment, the sampling arm must be calibrated in-situ to accommodate any distortions that occur during the multi-year space flight to the moon.  Moreover, these highly automated missions should be resilient to distortions of the robot and tool geometries that arise from unexpectedly forceful interactions with the terrain, joint freeze-ups in the $50^o K$ surface temperature, and improper loading of the sampling tools.  Practically speaking, icy moon missions require an automated calibration process that (1) provides sufficient accuracy, especially for the workspace regions where critical science occurs; (2) requires very few samples of arm data to reach a high recalibration accuracy, since in-situ recalibration consumes valuable mission time that should otherwise be devoted to science; and (3) can handle miscalibrations that are larger than typically experienced by industrial robots.


\begin{figure}[t]
	\centering
	\includegraphics[scale=0.4]{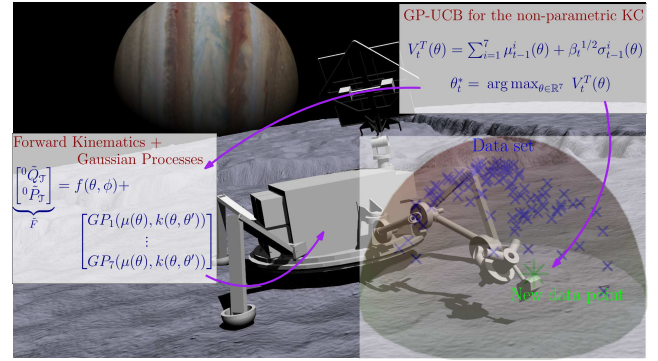}
	\caption{Visualization of the proposed active learning-based kinematic calibration framework on a space robotic arm. 
                 }
	\label{fig:scheme}
 \vskip -6 mm
\end{figure}

This paper develops an approach to manipulator kinematic calibration based on Gaussian Processes (GP) \cite{GP2006}. Gaussian Process Regression (GPR) can be used to learn kinematic residual errors, and allows for arm recalibration after large arm deformations.  More importantly, Gaussian Process Optimization \cite{srinivas2009} adaptively and autonomously selects a small set of arm configurations that provide accurate recalibration.  We demonstrate the utility of the proposed method on a 2-DOF planar manipulator, a 7-DOF Barrett WAM robot arm, 
and 6-DOF robot arm used in the OceanWATERS testbed \cite{catanoso2021}.

This paper is organized as follows. After some preliminaries in Sect. II, Sect. III introduces an active learning-based kinematic calibration framework.  Simulations are presented in Sect. IV, while Sect. V concludes the paper.


\vskip 10 pt
\noindent
{\bf Related Work:}
%
Kinematic calibration of a robot manipulator
has been extensively studied. Typically, the calibration process requires expert inputs for experiment design and numerous measurements to optimize an observability index. After data acquisition, offline optimization-based techniques, such as iterative least squares regression with linearized kinematic equations \cite{hollerbach1996}, Quadratic-Programming (QP) \cite{cursi2021}, and the Levenberg-Marquardt algorithm \cite{lembono2019} are used to identify the kinematic parameters.  Although these approaches improve robot positioning accuracy, they are neither data-efficient or capable of completely autonomous implementation. Both characteristics are needed for a deep-space application.

Machine learning-based frameworks can reduce positioning errors by considering issues, such as joint elasticities and backlash, that affect calibration accuracy. A Product-Of-Exponential kinematic model \cite{park1994} has been combined with a GP to compensate for residual deflections \cite{jing2016}.  In \cite{cursi2022}, an Artificial Neural Network directly learns robot position and orientation without nominal forward kinematic equations.  The post-calibration accuracy of these methods highly depends upon the quality of the input data, whose selection relies upon the calibration operator’s knowledge.  An unsupervised learning-based recalibration, which also optimizes the kinematic parameters for manipulator visual servoing, is proposed in \cite{gothoskar2022}.

Many previous works use observability indexes \cite{hollerbach1996} to optimize the choice of sampling locations.   An experimental comparison of the indexes is studied in \cite{joubair2013}. An active calibration algorithm is developed in \cite{sun2008} to find the most efficient set of calibration poses for the linearization-based techniques. A convex optimization approach to choosing configurations from a pool of candidate data points is proposed in \cite{kamali2019}.  The calibration experiment design methods in \cite{wu2015geometric, wang2017finding}, based on partial pose measurement, is performed before the experiments.   Although these methods are efficient for linearization-based calibration, they do not consider how to choose sampling points in the presence of larger inaccuracies. In \cite{du2020online}, an online calibration approach combines an unscented Kalman filter and iterative particle filter with a probabilistic sampling strategy. A Kalman filter-based online calibration method is presented in \cite{yang2022online} for legged robots.

\section{Preliminaries}
\vskip -1mm
We use the standard notation:  $\mathbb{R}$ and $\mathbb{R}^+$ represent the set of real and positive real numbers, respectively. The Euclidean norm of a matrix is denoted by $\|\cdot\|$.
$f \sim GP (\mu, k)$ denotes that the function $f(\cdot)$ is sampled from a Gaussian Process (which is reviewed in section \ref{section:GP}).

\subsection{Forward Kinematics}
\vskip -1mm
A forward kinematic function maps a robot's joint space to its workspace, and it is a crucial part of closed-loop kinematic motion analysis.  We adopt the widely used D-H convention that defines the relative position and orientation of the robot links and end effector using a four-parameter representation.\footnote{The methods of this paper can be readily adapted to a product-of-exponentials representation of manipulator kinematics.}  Given an $n$ jointed robot and the D-H parameters for the $i^{th}$ link: joint angle $\theta_i$, twist angle $\alpha_i$, link length $a_i$, link offset $d_i$, we define the transformation matrix ${^{i-1}T_i \in SE(3)}$, which consists of a rotation matrix ${^{i-1}R_i \in SO(3)}$ and a translation vector ${^{i-1}P_i \in \mathbb{R}^{3}}$, between neighboring robot links, {link~${i-1}$} and {link~${i}$}, as
\begin{align}
\label{Ti-1i}
&^{i-1}T_i  =  
\left[
\begin{array}{c|c}
^{i-1}R_i & ^{i-1}P_i \\
\cline{1-2}
0~~0~~0 & 1
\end{array}
\right] = \nonumber \\
&\begin{bmatrix}
\cos{\theta_i} & -\sin{\theta_i} \cos{\alpha_i} & \sin{\theta_i} \sin{\alpha_i} & a_i \cos{\theta_i}\\ 
\sin{\theta_i} & \cos{\theta_i} \cos{\alpha_i} & -\cos{\theta_i} \sin{\alpha_i} & a_i \sin{\theta_i}\\
0 & \sin{\alpha_i} &  \cos{\alpha_i} & d_i \\
0 & 0 &  0 & 1
\end{bmatrix}
\end{align}
for $i = 1,~2 \dots,~n$, where $\theta_i$ is variable if the joint is revolute, while $d_i$ is variable if the joint is prismatic. 
Then, the forward kinematics function is given by 
\begin{equation}
\label{T0T}
^{0}T_\mathcal{T} =~^{0}T_1(\theta_1, \phi_1) ^{1}T_2(\theta_2, \phi_2) \ldots  ^{n-1}T_{n}(\theta_n, \phi_n) ^{n}T_{\mathcal{T}},
\end{equation}
where $ {\theta =  \left[  \theta _1~ \ldots~ \theta _n \right]^T \in \mathbb{R}^{n}}$, $ {\phi =  \left[  \alpha _i~ d_i~ a_i \right]^T \in \mathbb{R}^{3n}}$, and $\mathcal{T}$ represents the tool frame. This function gives the orientation and position of the tool frame with respect to the base frame, i.e., $^{0}R_\mathcal{T} \in SO(3), ~^{0}P_\mathcal{T} \in \mathbb{R}^{3}$, and can be written more compactly as
  \begin{equation} \label{RP}
    \underbrace{ \begin{bmatrix}^{0}Q_\mathcal{T} & ^{0}P_\mathcal{T}
    \end{bmatrix}^T}_{F} = f(\theta, \phi),
  \end{equation}
where $^{0}Q_\mathcal{T}$ is a four-parameter quaternion representation of the rotation matrix \cite{murray2017}. 
Note that the quaternion conversion from a rotation matrix may be discontinuous, and therefore complex to model by Gaussian Processes\cite{zhou2019continuity}.  To circumvent this problem, we utilize the continuous quaternion conversion framework proposed in \cite{wu2019optimal}. 

\subsection{Kinematic Calibration Problems}
We now formalize two different calibration problems with respect to the kinematic uncertainty structure.  In the following definitions, let  $\tilde{F} =~[ ^{0}\tilde{Q}{}_\mathcal{T}~ ^{0}\tilde{P}{}_\mathcal{T} ]^T$ represent the actual (uncertain, after distortion) kinematic function. 

\vskip 8 pt
\noindent {\bf The Parametric calibration problem:} Consider a perturbed D-H convention parameter set
\begin{equation}
    [\theta ~ \phi]^T =  \left[ \theta_i + \Delta \theta_i ~ \alpha_i + \Delta \alpha_i  d_i + \Delta d_i ~ a_i + \Delta a_i \right]^T ,
\end{equation}
\begin{equation}
 \Delta =  \left[ \Delta \theta_i ~  \Delta \alpha_i ~ \Delta d_i ~ \Delta a_i \right]^T \in \mathbb{R}^{4n},
\end{equation}
 where $\Delta$ is the difference between the nominal (idealized) and actual D-H parameters. The parametric calibration problem is 
\begin{argmini}|s|
{[\phi]^T \in \mathbb{R}^{4n}}{\| \tilde{F}-F \|^2 }
{\label{Param}}
{ [\phi^*]^T =}
\end{argmini} 

\noindent {\bf The Non-parametric calibration problem:} Consider the nominal forward kinematic function $f_n(\theta, \phi)$ with an unstructured additive uncertainty term $ \Delta f(\theta, \phi)$ such that $f(\theta, \phi) = f_n(\theta, \phi) + \Delta f(\theta, \phi) $.
Then, the non-parametric calibration problem is defined as
\begin{argmini}
{\Delta f(\theta)}{\| \tilde{F}-F \|^2 }
{\label{nParam}}
{ \Delta f^* (\theta) =}
\end{argmini}
\noindent  That is, we want to find the {\em residual} function $\Delta f^*(\theta)$ that minimizes the calibration error with respect to the forward kinematic function that uses idealized D-H parameters.

Kinematic calibration methods are driven by error between the measured end-effector data and the computed forward kinematics data.
Therefore, an expert proposes enough experiments to provide sets of input joint angles $\theta$ and measurements $\tilde F$ to solve the given optimal error reduction problem.   Then, the objective function of the optimization problems in (\ref{Param}), (\ref{nParam}) is redefined on a discrete set of data as 
\begin{equation}
\label{OptMea}
 \Hat{\Delta}  F= 
 \begin{bmatrix}
 \tilde{F}_1 - F_1 &
  \ldots &
  \tilde{F}_m - F_m
\end{bmatrix}^T~\in \mathbb{R}^{7m},
\end{equation}
where $m$ is the number of measurements.


\subsection{Classical Linearized Recalibration}

The calibration cost function given in (\ref{Param}) is generally nonlinear and non-convex depending on the robot kinematic structure.  Thus, most classical recalibration techniques are based on a linearized version of the parametric calibration problem. By assuming relatively small parametric deflections, one can linearize the calibration function (\ref{RP}) as
\begin{equation}
\label{Lin}
 \underbrace{\tilde{F}-F}_{\Delta F \in \mathbb{R}^{7}} = 
\underbrace{ \begin{bmatrix}
\dfrac{\partial f}{\partial \alpha_i} & \dfrac{\partial f}{\partial d_i} & \dfrac{\partial f}{\partial a_i}    
\end{bmatrix}}_{J_{\phi} \in \mathbb{R}^{7 \times 3n}}
\underbrace{ \begin{bmatrix}
\Delta \alpha _i & \Delta d_i & \Delta a_i    
\end{bmatrix}^T }_{\Delta{\phi} \in \mathbb{R}^{3n}} ,
\end{equation}
where $J_{\phi}$ is the Jacobian matrix.
Therefore, when the nominal D-H parameters and their possible lower bound ${\Delta{\phi}_{lb} \in \mathbb{R}^{3n}}$ and upper bound ${\Delta{\phi}_{ub} \in \mathbb{R}^{3n}}$ are given, the linearization-based KC problem can be solved via the following constrained QP:
\begin{argmini*}|s|
{\Delta{\phi} \in \mathbb{R}^{3n}}{\| \tilde{F} - F - J_{\phi} \Delta{\phi} \|^2 }
{\label{Lin-QP}}
{ \Delta{\phi}^* =}
\addConstraint{\Delta{\phi}_{lb} \leq \Delta{\phi}  \leq \Delta{\phi}_{ub} }
\end{argmini*}
Then, we obtain $\phi^* = \phi + \Delta{\phi}^*$. This QP is iteratively solved until an acceptable error convergence is achieved \cite{cursi2021}.

\subsection{Gaussian Processes (GPs)}
\label{section:GP}
Gaussian Processes provide a stochastic, data-driven, supervised machine learning approach to specify the relations between input and output data sets of smooth functions through Bayesian inference \cite{GP2006}. A GP function, ${f(x): X \xrightarrow[]{} \mathbb{R}}$, defined on a data domain $X$, is fully specified by its mean function ${\mu (x): X \xrightarrow[]{} \mathbb{R}}$ and a covariance/kernel function ${k(x, x'): X \times X \xrightarrow[]{} \mathbb{R}}$:
\begin{align}
\label{GPDef}
     \mu(x) & \triangleq \mathop{{}\mathbb{E}} \left[ f(x) \right],\\
    \label{GP1}
     k(x, x') & \triangleq \mathop{{}\mathbb{E}} \left[ (f(x) - \mu(x) ) (f(x') - \mu(x') ) \right].
\end{align}
We say that $f(\cdot)$ is distributed according to:
  \begin{equation} \label{GP2} f(x)  \sim GP (\mu(x), k(x, x')). 
  \end{equation}
  
Given a collection of training data $\mathcal{D} \triangleq \left\{ X_N \in X, ~ Y_N \in \mathbb{R}^{N} \right\}$ with inputs $X_N = \left [ x_1, \ldots , x_N \right ] $, outputs $Y_N = \left [ f(x_1) + \epsilon_1, \ldots , f(x_N)+ \epsilon_N \right ]^T $ and i.i.d zero-mean Gaussian noises ${\epsilon_i \sim \mathcal{N} (0, \sigma^2_{\epsilon})}$, a GP predicts a set of outputs based on an input test data set ${ \tilde{X} = \left [ \tilde{x}_1, \ldots , \tilde{x}_M \right ] \in X}$. The posterior of the GP over the observations, $\tilde{f} \big | \tilde{X}, \mathcal{D}$, 
is also a GP distribution with a mean $\tilde{\mu}$ and variance $\tilde{\sigma}^2$ given by
\begin{align}
\label{deGP0}
\tilde{f} & \big | \tilde{X}, \mathcal{D}   \sim \mathcal{N} \big ( \tilde{\mu}, \tilde{\sigma}^2 \big ) ,  \\
\label{deGP1}
  {\tilde{\mu}} &=  K_{\tilde{X} X_N}  (K_{X_N X_N} + \sigma^2_{\epsilon} I )^{-1}~Y_N, \\
  \label{deGP2}
   {\tilde{\sigma}^2} &= K_{\tilde{X} \tilde{X}} -  K_{\tilde{X} {X}_N} ( K_{X_N X_N} + \sigma^2_{\epsilon} I )^{-1} K_{{X}_N \tilde{X} },
\end{align}
where $I$ is the identity matrix, ${K_{X_N X_N} \in \mathbb{R}^{N \times N}}$, ${K_{\tilde{X} \tilde{X}} \in \mathbb{R}^{M \times M}}$, ${K_{\tilde{X} X_N} \in \mathbb{R}^{M \times N}}$, ${K_{ X_N \tilde{X}} \in \mathbb{R}^{N \times M}}$ are the covariance matrices for the set of points, which measure the correlations between the inputs with ${ [K_{X_N X_N}]_{i, j} = k(x_i, x_j)}$, ${ [K_{\tilde{X} \tilde{X}}]_{i, j} = k(\tilde{x}_i, \tilde{x}_j)}$, ${ [K_{\tilde{X} X_N}]_{i, j} = k(\tilde{x}_i, {x}_j)}$, ${ [K_{X_N \tilde{X}}]_{i, j} = k({x}_i, \tilde{x}_j)}$.

Briefly, the kernel function reflect the features of the functions to be learned, such as smoothness and input correlations.   Our experiments use the commonly used squared-exponential (SE) kernel given by
\begin{equation}
\label{SEK}
k(x_i, x_j) = \sigma^2_{f}~exp \bigg (  \dfrac{-\|x_i - x_j\|^2}{2 l^2}  \bigg ) + \sigma^2_{n} \delta(x_i, x_j) ,
\end{equation}
where ${\theta_h = [ l~\sigma_{f}~\sigma_{n} ]^T}$ are the tunable hyperparameters and $\delta(x_i, x_j)$ is the Kronecker delta function.
The optimal hyperparameters can be found by minimizing the negative log marginal likelihood function \cite{GP2006}.  For data set $\mathcal{D} $, this function is defined as
\begin{align}
    \label{MLL}
    \nonumber
    & \log p (Y_N \big | X_N , \theta_h) = -\dfrac{1 }{2} Y_N^T (K_{X_N X_N}(\theta_h) + \sigma^2_{\epsilon} I )^{-1} Y_N \\
    & -\dfrac{1 }{2} \log \big | (K_{X_N X_N}(\theta_h) + \sigma^2_{\epsilon} I ) \big | - \dfrac{N }{2} \log 2 \pi .
\end{align}
Finally, the optimal hyperparameters $\theta_h^*$, and $\sigma_{\epsilon}^*$ can be computed via the gradient-descent based solution of the following typically non-convex optimization problem:
\begin{argmini}|s|
{\theta_h,\sigma_{\epsilon}}{-\log p (Y_N \big | X_N , \theta_h). }
{\label{MLLParam}}
{ [ \theta_h^*~ \sigma_{\epsilon}^* ]^T =} 
\end{argmini} 

\section{Active Learning Based Kinematic Calibration Approach}
\vskip -0 mm
\subsection{GPs for Non-parametric Calibration}
\label{section3a}
In this section, the residuals representing the inaccuracies in the forward kinematics are modelled via GPs. Consequently, we will be able to achieve effective non-parametric robot kinematic learning and calibration using noisy sensor measurements in a stochastic framework.

The kinematic residual errors can be captured with seven GPs along each orientation and translation direction.  In particular, each component of matrix $F$ in (\ref{RP}) is modelled as a single-output GP with a mean and kernel function as
\begin{equation}
\label{RP4KC}
\underbrace{ \begin{bmatrix}
^{0}\tilde{Q}_\mathcal{T} \\
^{0}\tilde{P}_\mathcal{T}
\end{bmatrix} }_{\tilde{F}}
= f(\theta, \phi) +
\begin{bmatrix}
GP_1 (\mu(\theta), k(\theta, \theta')) \\
\vdots \\
GP_7 (\mu(\theta), k(\theta, \theta')) 
\end{bmatrix} ,
\end{equation}
where $\theta$ is the vector of joint angles, and $\tilde{F}$ is the measured tool-frame poses.  The regressed GP functions  can predict the residual error and its variance in each direction of motion, at a new query input $\theta^*$, by using (\ref{deGP1}) and (\ref{deGP2}).

Our GP-based modeling of the residual errors in each motion direction is simplistic. One should expect coupling in the errors across these directions.  Multi-output Gaussian Processes (MOGPs) \cite{bruinsma2020} and vector-valued GPs \cite{hutchinson2021} are suitable solutions to this problem. However, due to the computational scaling of these methods, we assume single-output GPs. As shown in Section~\ref{Experiments},
the Gaussian distribution property holds in our applications.  Thus, the decoupled GP assumption is reasonable and computationally efficient.

\subsection{GPs for Parametric Calibration}
\vskip -0 mm
We briefly note that GP regression can be used for  parametric calibration, all input parameters of the kinematic function in (\ref{RP}) can be modelled as a single-output GP. To this end, we modify (\ref{RP}) as
  \begin{equation} \label{RPinv}
      \begin{bmatrix}\theta & \phi
    \end{bmatrix}^T
    = f^{-1}(^{0}Q_\mathcal{T},^{0}P_\mathcal{T}) ,
  \end{equation}
where the D-H parameters, including joint angles, are outputs, and the tool frame poses are inputs. By assuming that each component in ${[\theta ~ \phi]^T \in \mathbb{R}^{4n}}$ can be modeled by a GP, the parametric calibration problem can be written as
\begin{equation}
\label{RP4PKC}
\begin{bmatrix}
\tilde{\theta} \\ \tilde{\phi}
\end{bmatrix} 
=
\begin{bmatrix}
{\theta_0} \\ {\phi_0}
\end{bmatrix} 
+
\begin{bmatrix}
GP_1 (\mu(F), k(F, F')) \\
\vdots \\
GP_{4n} (\mu(F), k(F, F')) 
\end{bmatrix} ,
\end{equation}
where ${\tilde{\theta},~\tilde{\phi},~F}$ are the measurements, ${{\theta_0},~{\phi_0}}$ are the nominal values of the D-H parameters.

As compared to the non-parametric approach in (\ref{RP4KC}), the joint angle vector is also assumed to be uncertain. Therefore, Eq.~(\ref{RP4PKC}) can model the possible joint angle offsets. 
But, the number of GPs needed in this approach increases with the number of robot joints. For example, for a 7 DOF robot, 28 different GPs must be defined. Because of these drawbacks, we leave the parametric approach to our future studies.

\subsection{GP-UCB for Non-parametric Experiment Design}
\vskip -0 mm
We are concerned not only with fitting a GP model to calibration data, but also with designing a calibration experiment that minimizes the number of samples.  such optimization is a classical experimental design problem that aims to find the best set of measurement points \cite{sun2008active, hollerbach2016model}.  When using  GP regression-based calibration alone, a user must select the training and test data points, which are essential for the method's accuracy.  This section adopts a Gaussian Process Upper Confidence Bound (GP-UCB) \cite{srinivas2009} active learning algorithm to adaptively select a sequence of informative measurements to minimize the residual error.

Modeling the residual kinematic error with a GP, as in Section \ref{section3a}, brings a significant advantage for experiment design, since it encodes residual function uncertainty in the confidence bounds \cite{schulz2018}. With this feature, the GP-UCB algorithm uses currently available measurements and the residual function confidence bounds to decide the next test point that optimizes the utility function $V_t(x): X \to \mathbb{R}$
\begin{align}
\label{utility}
    {V_t(x) = \ } & {\mu_{t-1}(x)} + {\beta_t}^{1/2} \sigma_{t-1}(x) , \\
\label{utilityopt}
    {x_t^*= \ } &
\operatorname*{arg\,max}_{x \in X}~  V_t(x) ,
\end{align}
where $\beta_t \in \mathbb{R}^+ $ is an iteration-varying parameter \cite{srinivas2009} that weights residual function uncertainty in the selection of the next sampling point, $\sigma_{t-1}(x)$ is the square root of the variance, and $\mu_{t-1}(x)$ is the predicted value of the mean function at input $x$.
Hence, GP-UCB chooses sampling points that trade off exploitation of joint space regions with high calibration performance against exploration of regions with high uncertainty about the residual function.  

The GP-UCB algorithm has an appealing {\em no regret} property \cite{srinivas2009}.  I.e., the algorithm is {\em guaranteed} to find the globally optimal residual functions given sufficient examples. Our experiments below show that some other calibration experiment design approaches become stuck in local minima.  Moreover, its {\em instantaneous regret} is proportional to $T^{-1/2}$, where $T$ is the number of experiments.  This property implies that the calibration error steadily decreases with each sample.  While this paper focuses on sequential sampling selection, a variation of this approach, entitled GP-BUCB \cite{desautels2014}, proposes {\em batches} of experiments, while still retaining a no-regret property and similar instantaneous regret performance.  

Since there are seven different GPs with the same input vector, we modify the utility function as the sum of UCBs:  
\begin{equation}
\label{utilitymod}
    V_t^T(x) = \sum_{i=1}^{7} {\mu_{t-1}^i (x)} + {\beta_t}^{1/2} \sigma_{t-1}^i (x) ,
\end{equation}
where ${\mu_{t-1}^i (x)},~\sigma_{t-1}^i (x)$ are the mean and variance of the $i^{th}$ output given in (\ref{RP4KC}). Hence, the resulting new putatively optimal data sampling point is obtained by considering all GPs simultaneously. The authors in \cite{shen2022} show that this approximation does not violate the no-regret property of the UCB algorithm under mild conditions.

The basic pseudo-code of the proposed GP-UCB-based calibration approach is presented in Algorithm~\ref{alg1}, which starts with any available GP prior, and the next measurement is chosen to maximize the utility function  (\ref{utilitymod}).
\begin{algorithm}[b]
 \caption{GP-UCB for non-parametric arm calibration}
 \label{alg1}
 \begin{algorithmic}[10]
 \renewcommand{\algorithmicrequire}{\textbf{Input:}}
 \renewcommand{\algorithmicensure}{\textbf{Output:}}
 \REQUIRE Input set $\theta \in X$ \\~~~~GP prior $\mu_0,~\sigma_{0},~k$ \\~~~
 Kernel function $k(\cdot,\cdot)$ \\~~~
 Parameter $\beta_t$
 \ENSURE  $\theta^*$
  \FOR {$t = 1, 2, \ldots$,}
  \STATE Choose
$ {\theta_t^*= \ }
\operatorname*{arg\,max}_{\theta \in X}~ 
V_t^T(\theta)$
  \STATE Compute $\sigma_{t}$ via Equation (\ref{deGP2})
  \STATE Obtain ${\Delta F}_t = ( \tilde{F}(\theta_t) - F(\theta_t)) + \epsilon_t $  
  \STATE Compute $\mu_{t}$ via Equation (\ref{deGP1})
  \ENDFOR
 \end{algorithmic} 
 \end{algorithm}
The GP residual is updated from the new measurement (using Eqs. (\ref{deGP1}) and (\ref{deGP2})).   This iteration continues until a stopping criterion is satisfied.  For the algorithm parameter ${\beta_t }$, which is defined in Theorem 2 of \cite{srinivas2009}, we use a strategy proposed in \cite{srinivas2009} as
\begin{align}
\label{beta}
    \beta_t = & 2 log (t^2 2 \pi^2 / (3 \delta_{\beta}))  ~+ \nonumber \\ & 2 d_{\beta} log \big (t^2 d_{\beta} a_{\beta} b_{\beta} r_{\beta} \sqrt{log(4d_{\beta}/\delta_{\beta})}\big  ).
\end{align}
\vskip -0 mm
\begin{figure*}[t]
\centering
  \begin{subfigure}{5.0cm}
    \centering\includegraphics[width=5.0cm]{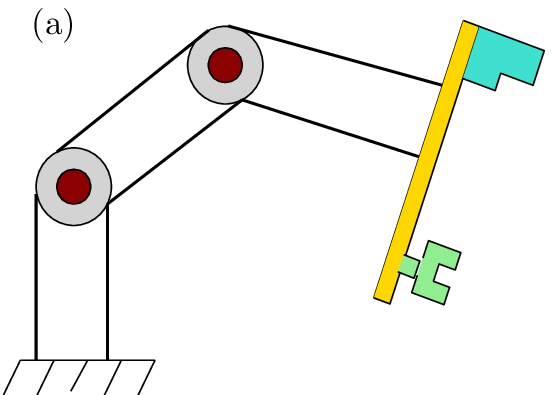}
  \end{subfigure} 
   \begin{subfigure}{5.2cm}
    \centering\includegraphics[width=5.2cm]{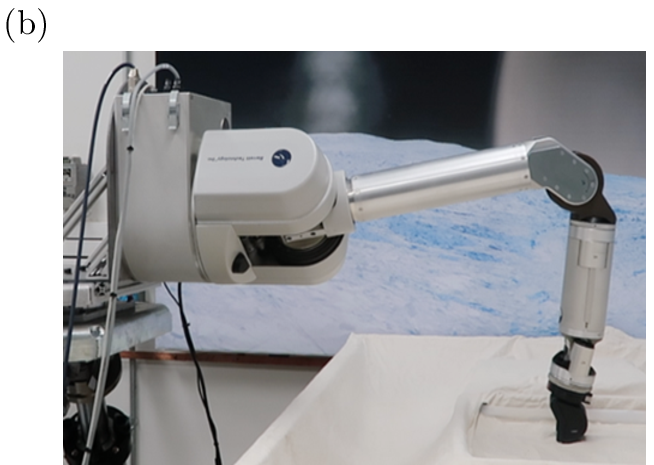}
  \end{subfigure} 
  \begin{subfigure}{5.3cm}
    \centering\includegraphics[width=5.3cm]{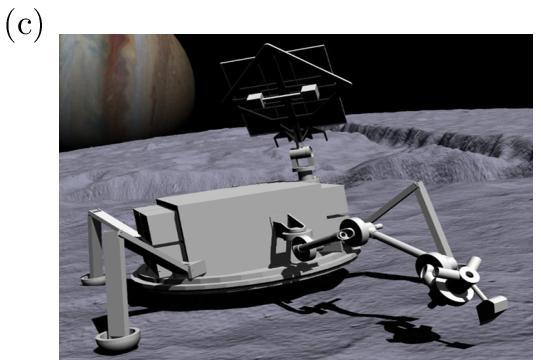}
  \end{subfigure} \\ [0.2cm]
\begin{subfigure}{5.2cm}
    \centering\includegraphics[width=5.2cm]{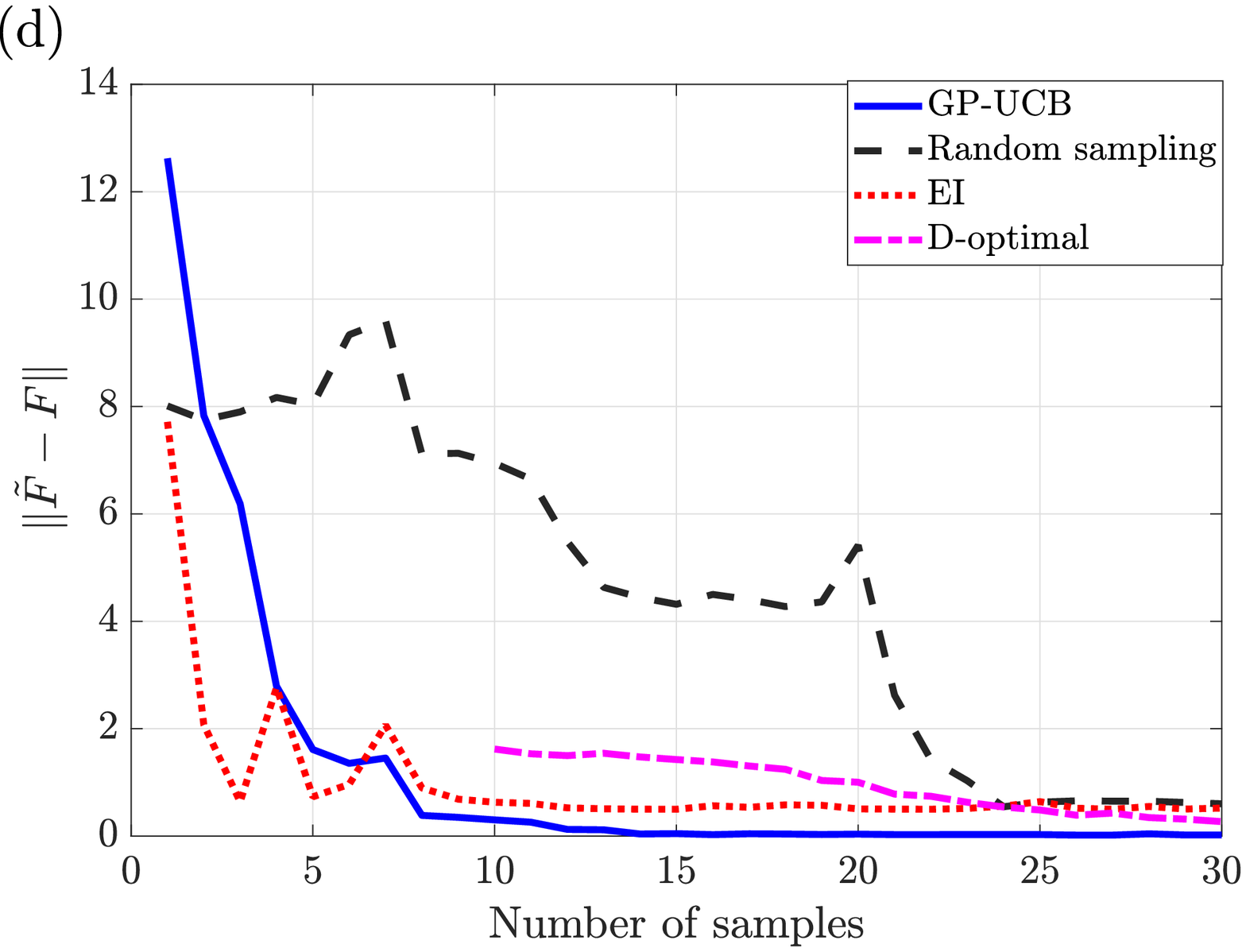}
\end{subfigure}
\begin{subfigure}{5.2cm}
    \centering\includegraphics[width=5.2cm]{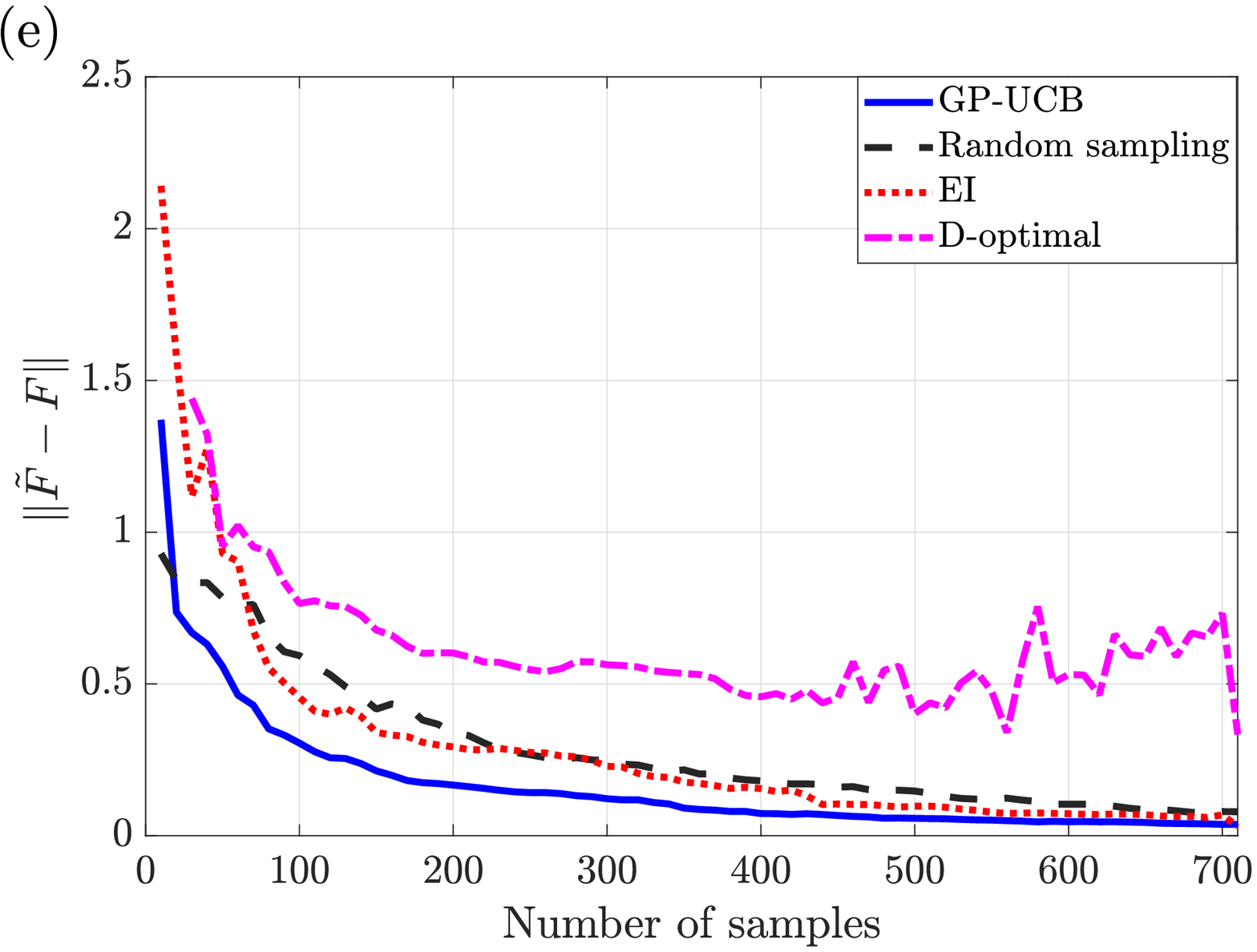}
  \end{subfigure}
\begin{subfigure}{5.2cm}
    \centering\includegraphics[width=5.2cm]{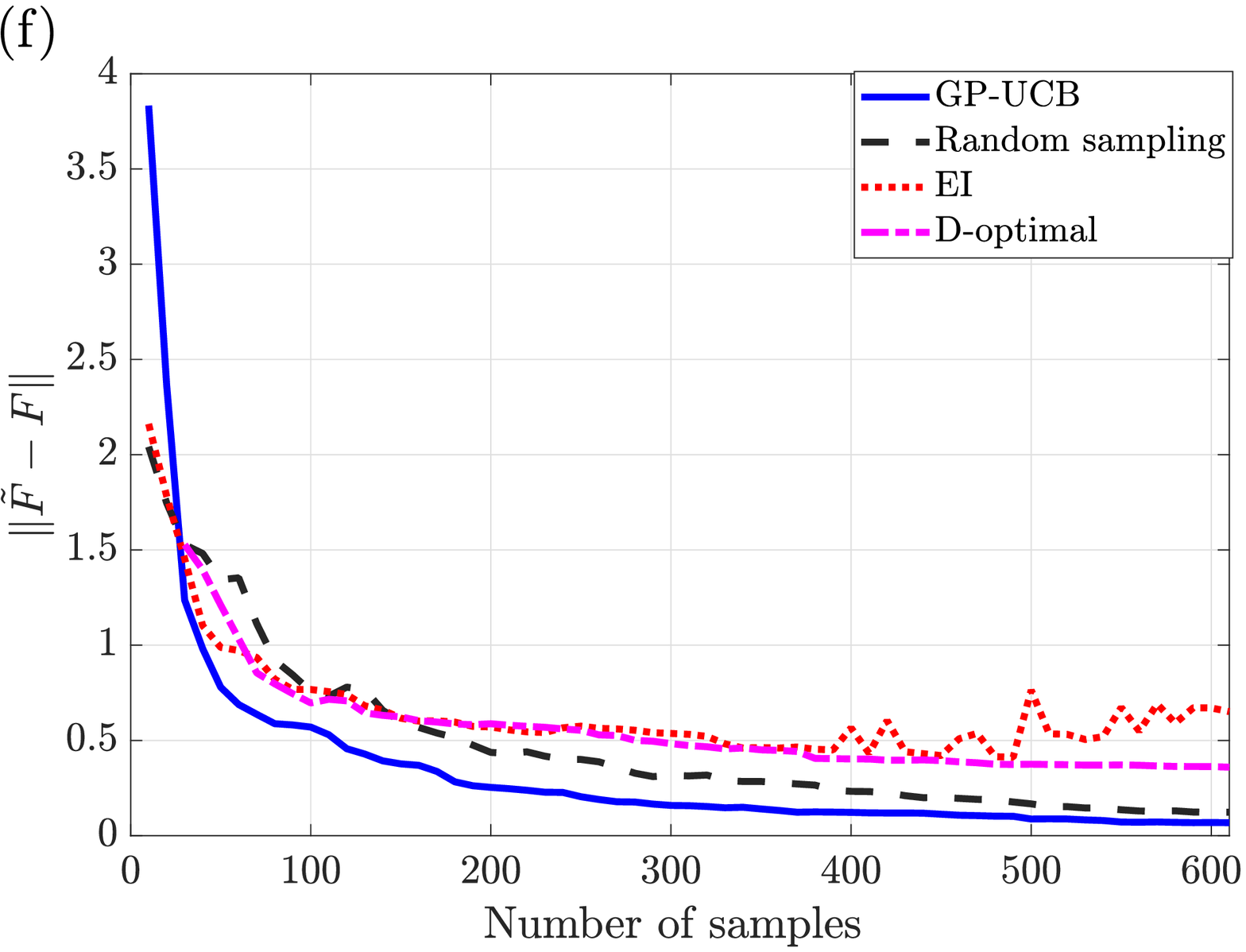}
  \end{subfigure} \\ [0.2cm]
  \begin{subfigure}{5.2cm}
    \centering\includegraphics[width=5.2cm]{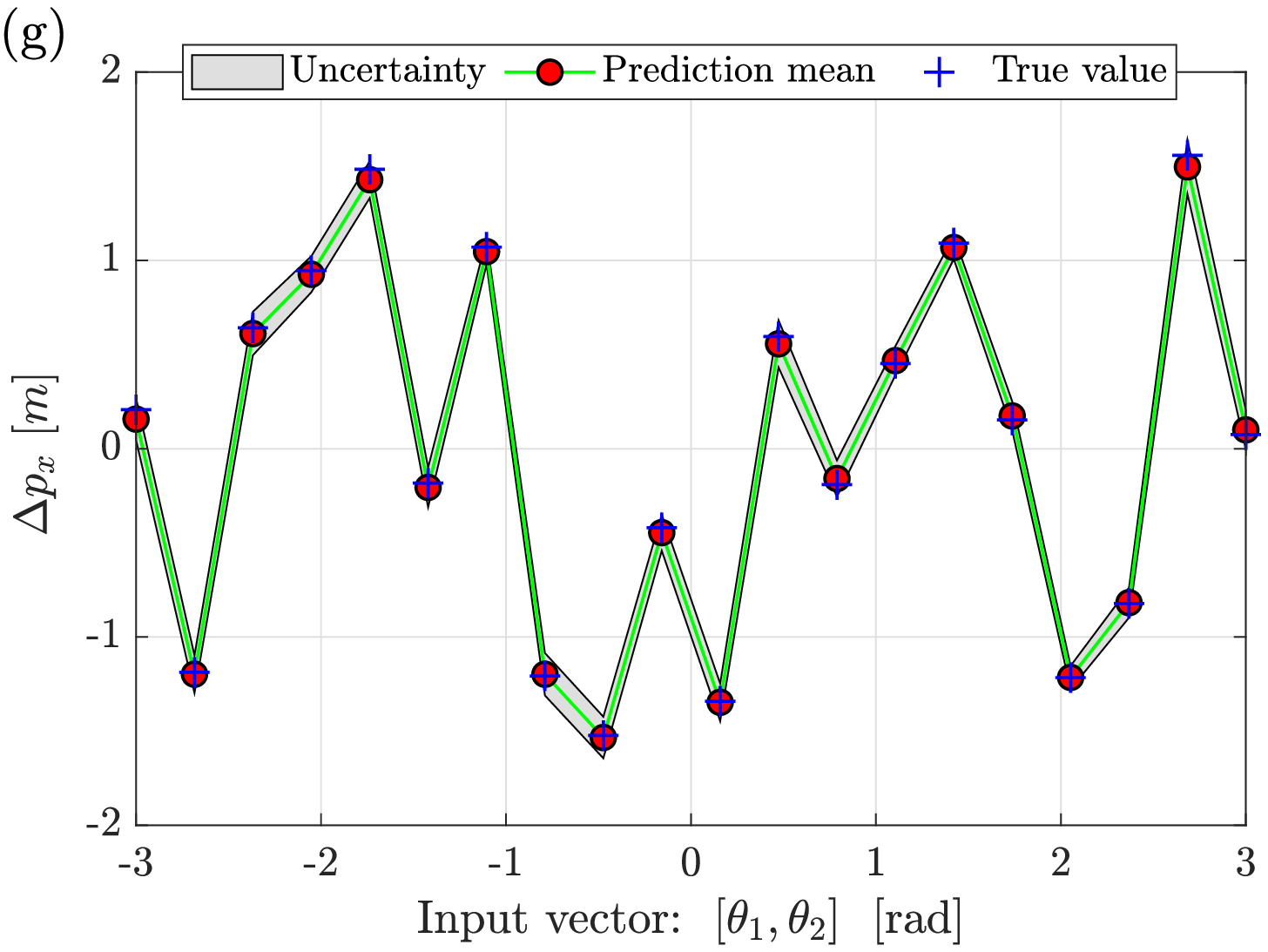}
  \end{subfigure}
  \begin{subfigure}{5.2cm}
    \centering\includegraphics[width=5.2cm]{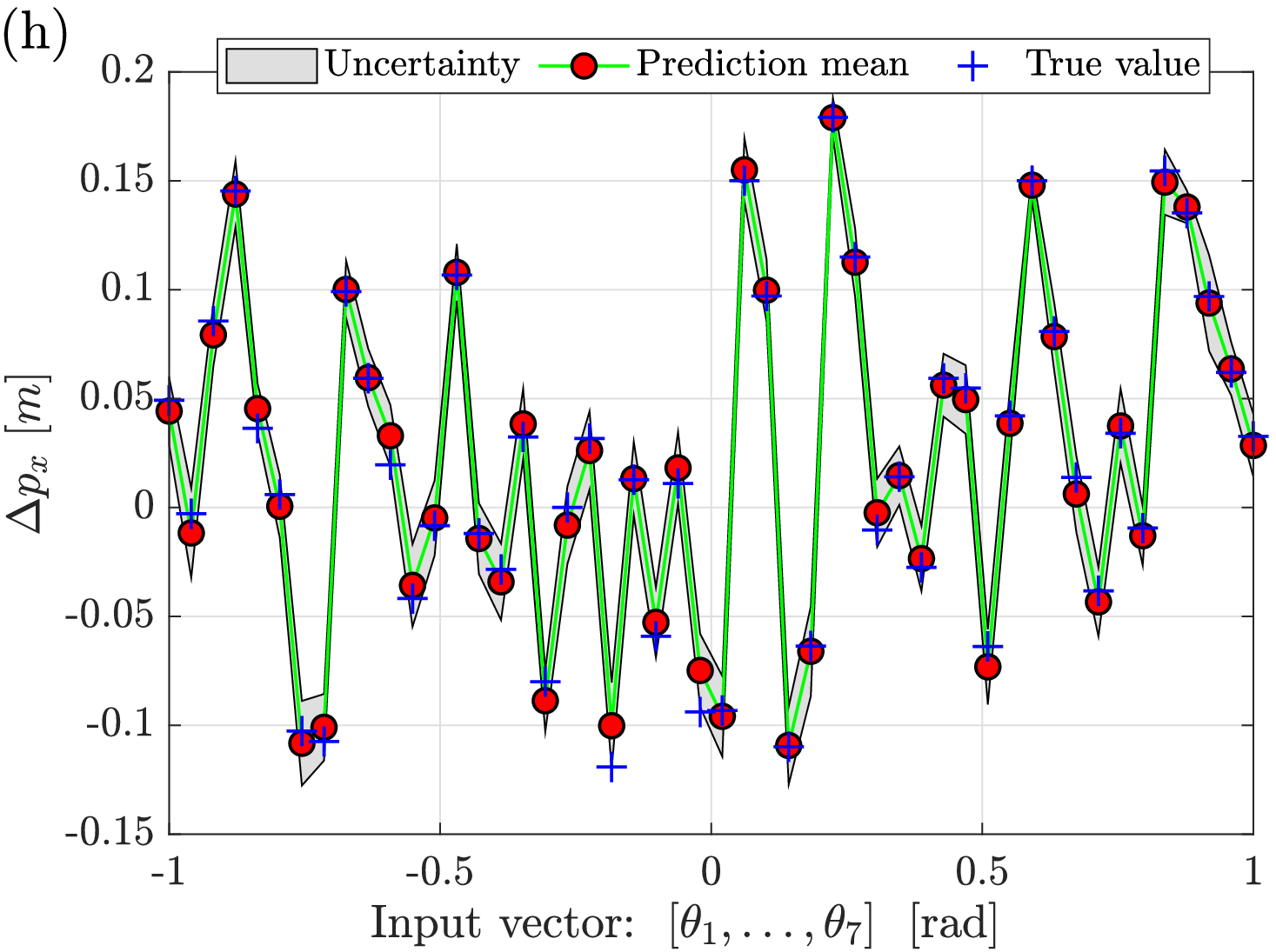}
  \end{subfigure}
   \begin{subfigure}{5.2cm}
    \centering\includegraphics[width=5.2cm]{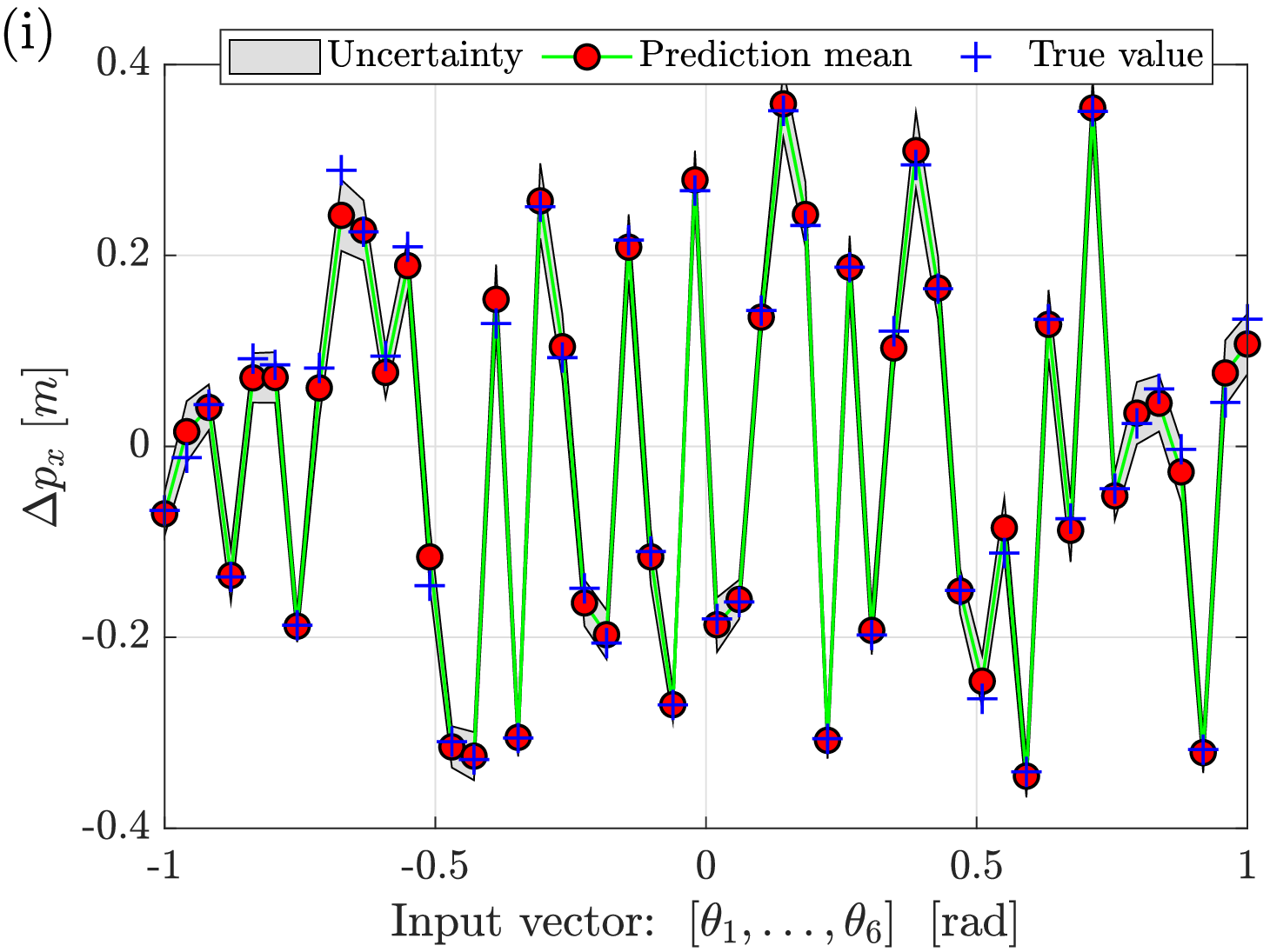}
  \end{subfigure}
  \vskip -0.0mm
  \caption{Simulation examples. (a) 2-DOF Manipulator. (b) 7-DOF Barrett WAM manipulator arm in OWLAT. (c) 6-DOF OceanWATERS Lander Robot Arm. (d), (e), (f) The evaluations of the calibration errors with respect to the number of iterations for 2-DOF, 7-DOF, and 6-DOF robots, respectively. (g), (h), (i) The errors in position $x$ obtained from the test samples for 2-DOF, 7-DOF, and 6-DOF robots, respectively.}
  \label{robots}

\end{figure*}

\section{Experiments} 
\label{Experiments}
 \vskip -0 mm
We evaluate the performance of the GP-based residual regression and GP-UCB-based sampling selection in three simulated robotic arms with perturbed kinematic parameters.
We compare our results with three other techniques: 1) A griding-based {\em D-optimal design algorithm} with a convex approximation proposed in \cite{kamali2019}. This algorithm utilizes a pool of possible joint angle sets. Then, the D-optimality criteria is adopted using the Jacobian matrix $J_{\phi}$ given in (\ref{Lin}) and this pool.  In \cite{kamali2019}, it is shown that D-optimal experimental design leads to better performance than random sampling and the DEMAX algorithm.  2) {\em Expected improvement (EI)} based sampling algorithm \cite{shahriari2015taking}. 
Like the UCB utility function in (\ref{utilitymod}), we modify the EI algorithm's utility to sum the expected improvements in each direction.
3) {\em Random sampling}.
 
\subsection{2-DOF Manipulator}
 \vskip -0 mm
We start with an easy-to-understand simulated calibration of a 2R planar robot (Fig.~\ref{robots}-(a)).  We seek to learn the forward kinematic residuals while choosing a small set of sampling points.  The uncertain D-H parameters of the arm are $d_1 = d_2 = 0 \pm 0.1 ~m$, $a_1 = a_2 = 1 \pm 0.2~m$, $\alpha_1 = \alpha_2 = 0 \pm 0.1~rad$, $ \theta_1, \theta_2 \in [-3,~ 3]~rad$.  
Even in this simple example, linearization-based methods may not yield an accurate calibration in the presence of additive uncertainties, 
since these approaches are formulated to mainly address constant inaccuracies, i.e., ${\phi = \phi_0 + e_{\phi}}$. 

We assume precise measurements of the joint angles {($\theta_1, \theta_2$)} and the orientation and position of tool frame {($F \in \mathbb{R}^{7}$)} (e.g., by a calibrated high-resolution camera).  We run Algorithm~\ref{alg1} for $t = 1, 2, \ldots, 30$ iterations, and choose the $\beta_t$ in (\ref{beta}) and an SE kernel (\ref{SEK}) with the hyperparameters optimized using~(\ref{MLLParam}) for all simulation examples.

Fig.~\ref{robots}-(d) plots the calibration error, $\|\tilde{F} - F\| $, versus the number of samples selected by GP-UCB. As more samples are collected, the calibration error norm converges to zero, confirming the no-regret property of this approach. EI-based sampling initially concentrates on the exploration of the calibration error functions. But after 10 steps, it encounters a local minimum, and does not continue to explore.  For the D-optimal sampling, we start with 10 randomly selected data points, since this method requires as many data points as the number of unknowns. Although this method initially offers better performance, the convergence rate is slow. While random sampling also provides high-precision kinematic error modelling, it requires significantly more sampling points than the active learning approach, as shown in Fig.~\ref{robots}-(d).  This confirms that GP-UCB is intelligently selecting its experiments. The post-calibrated performance of GP-UCB is shown in Fig.~\ref{robots}-(g) for the $x$ position error. 

\begin{figure}
	\centering
	\includegraphics[scale=0.32]{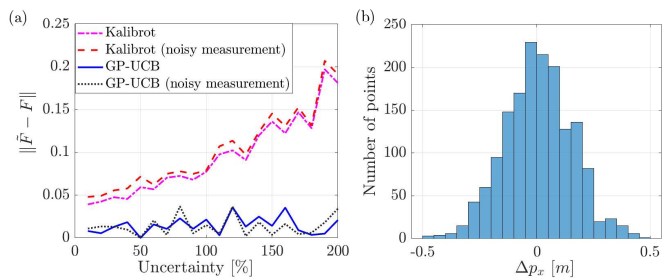}
	\caption{(a) Kalibrot vs GP-UCB. (b) Histograms of calibration error distribution along x-axis before the calibration process.}
	\label{fig:kalibrot}
 \vskip -0.0 mm
\end{figure}

To demonstrate our methods ability to handle larger kinematic errors, and noisy  measurements, the planar robot's D-H parameters are perturbed at a mean level starting from $10 \% $ up to to $200 \% $. Secondly, a normally distributed noise with a variance of 0.01 has been added to each output. For comparison, we used the Kalibrot toolbox \cite{cursi2021},  which relies on classical linearized recalibration.  Fig.~\ref{fig:kalibrot}-(a) shows that our method  learns and tolerates increasing kinematic uncertainties and is sufficiently robust to sensor noise. Although Kalibrot is robust against sensor noise, its calibration error increases roughly linearly with parameter uncertainty.

\subsection{7-DOF Barrett WAM Robot Arm}
 \vskip -0 mm
The OWLAT testbed was developed to evaluate autonomous techniques for \textit{Ocean Worlds} icy moon lander missions \cite{nayar2021}. A 7-DOF Barrett WAM manipulator simulates robotic sampling operations (Fig.~\ref{robots}-(b)) that might take place on an icy moon mission. The arm's kinematic parameters are given in Table~\ref{DH_7dof} with additive uncertainty bounds. 

Fig.~\ref{robots}-(e) shows the simulated convergence of the forward kinematic error estimate versus GP-UCB sample number.  
The results in this figure suggest that the calibration performance and data efficiency of our framework are better than other sampling methods for this 7-DOF arm, especially for small sample number. Although the EI algorithm gave the second best results for this robot arm, this method's lack of design parameters is a weakness. The gridding-based D-optimal algorithm fails in this simulation.  Fig.~\ref{robots}-(h) shows the post-calibrated kinematic accuracy on 50 test points linearly spaced along the $x$ position. 

\begin{table*}[t]
\centering
\begin{tabular}{|l|c|c|c|c|c|c|c|}
 \hline
 \multicolumn{8}{|c|}{\textbf{7-DOF Barrett WAM Robot Arm}} \\
 \hline
 Joint & $\pmb{1}$ & $\pmb{2}$ & $\pmb{3}$ & $\pmb{4}$ & $\pmb{5}$ & $\pmb{6}$ & $\pmb{7}$ \\
\hline
$\pmb{d}$ & $0 \pm 0.01$ & $0 \pm 0.02$ & $0.55 \pm 0.2 $ & $0 \pm 0.03$ & $0.3 \pm 0.2 $ & $0 \pm 0.04$ & $0.06 \pm 0.06$ \\
$\pmb{a}$ & $0  \pm 0.01$ & $0  \pm 0.03$ & $0.045 \pm 0.01 $ & $-0.045 \pm 0.01$ & $0 \pm 0.07$ & $0 \pm 0.1$ & $0 \pm 0.01$ \\
$\pmb{\alpha}$ & $-\pi/2  \pm 0.2 $ & $\pi/2 \pm 0.2 $ & $-\pi/2 \pm 0.3$ & $\pi/2 \pm 0.2 $ & $-\pi/2 \pm 0.1 $ & $\pi/2 \pm 0.1$ & $0 \pm 0.3 $ \\
$\pmb{\theta}$ & $\theta_1$ & $\theta_2$ & $\theta_3$ & $\theta_4$ & $\theta_5$ & $\theta_6$ & $\theta_7$ \\
\hline
 \hline
 \multicolumn{8}{|c|}{\textbf{6-DOF OceanWATERS Lander Robot Arm}} \\
 \hline
$\pmb{d}$ & $0 \pm 0.01$ & $0 \pm 0.02$ & $0 \pm 0.06 $ & $-0.15 \pm 0.01$ & $0 \pm 0.04 $ & $0 \pm 0.06$ & - \\
$\pmb{a}$ & $0.16  \pm 0.1$ & $0.37  \pm 0.1$ & $0.05 \pm 0.02 $ & $0.463 \pm 0.1$ & $-0.238 \pm 0.1$ & $0.225 \pm 0.02$ & - \\
$\pmb{\alpha}$ & $\pi/2  \pm 0.2 $ & $0 \pm 0.2 $ & $\pi \pm 0.3$ & $0 \pm 0.2 $ & $0 \pm 0.1 $ & $\pi/2 \pm 0.1$ & - \\
$\pmb{\theta}$ & $\theta_1$ & $\theta_2$ & $\theta_3$ & $\theta_4$ & $\theta_5$ & $\theta_6$ & - \\
\hline
\end{tabular}
\caption{DH parameters for the 7-DOF Barrett WAM Robot and 6-DOF OceanWATERS Lander Robot Arm.}
\label{DH_7dof}
\vskip -0.0 mm
\end{table*}

\subsection{6-DOF OceanWATERS Lander Robot Arm}
 \vskip -0 mm
We additionally demonstrate the proposed algorithm on the open-source OceanWATERS simulation testbed\cite{catanoso2021} (see Fig.~\ref{robots}-(c) and Fig.~\ref{fig:scheme}), which is a likely design for a future lander mission. The arm's nominal kinematic parameters and our tested uncertainty bounds are given in Table~\ref{DH_7dof}. 

Since our goal is to fit a GP to the residuals, we present the distribution of forward kinematic errors for 1500 sample points, chosen by the GP-UCB algorithm, in Fig.~\ref{fig:kalibrot}-(b). The figure demonstrates that the kinematic error has a nearly normal distribution with zero mean. This result confirms that the use of a Gauss Process framework is warranted.


Fig.~\ref{robots}-(f) compares the calibration performance of GP-UCB and the other approaches over 600 sampling iterations.
The figure shows that GP-UCB is an efficient sampling method for this robot, since fewer data points are needed for accurate recalibration.
The EI and D-optimal algorithms reduce the norm of the calibration error; but they become "stuck" on one sample  point after $\sim$300 samples. The post-calibration accuracy of the GP model is demonstrated on 50 linearly spaced $x$-axis test sample points in Fig.~\ref{robots}-(i).

\section{Conclusions and Future Work}
Motivated by the proposed use of manipulators in future icy moon sampling missions, this study aimed to accurately learn the manipulator forward kinematic error residual in an online, non-parametric, and data-efficient way.  GP regression models the kinematic inaccuracies and their confidences, 
while an online sequential GP-UCB algorithm efficiently selects the optimal measurement in a way that adapts to the specific errors experienced by the robot. 
Our numerical studies on three different arm geometries found that this framework generally outperformed other methods in the limit of a small number of calibration experiments, which is vital for deep space missions. 
In future work, we will investigate specific GP kernels that better capture the uncertainty properties of manipulator kinematics, and study the effects of measurement noise on calibration accuracy under GP and classical calibration frameworks. We will also evaluate if coupled Multivariate Gaussian Process models \cite{de2021mogptk,hutchinson2021,moriconi2020high} may further improve the calibration accuracy and sample efficiency.  

\bibliographystyle{IEEEtran}
\bibliography{References.bib}

\end{document}